%% file: main.tex
\documentclass[letterpaper]{article} 
\usepackage{aaai2026}  

\usepackage{times}  
\usepackage{helvet}  
\usepackage{courier}  
\usepackage[hyphens]{url}  
\usepackage{graphicx} 

\usepackage{lineno} 
\usepackage{pifont} 

\usepackage{natbib}  
\usepackage{caption} 
\usepackage{algorithm}
\usepackage[noend]{algpseudocode} 
\usepackage{newfloat}
\usepackage{multirow}

\usepackage{listings}
\DeclareCaptionStyle{ruled}{labelfont=normalfont,labelsep=colon,strut=off} 
\floatstyle{ruled}
\newfloat{listing}{tb}{lst}{}
\floatname{listing}{Listing}
\usepackage{newfloat}
\usepackage{listings}
\def\Ourmodel{RoomPlanner}

\DeclareCaptionStyle{ruled}{labelfont=normalfont,labelsep=colon,strut=off} 
\floatstyle{ruled}
\newfloat{listing}{tb}{lst}{}
\floatname{listing}{Listing}
\nocopyright
\usepackage{amsmath}
\usepackage{amsfonts}
\usepackage{bm}
\usepackage{cleveref}
\crefname{section}{Sec.}{Secs.}
\crefname{equation}{Eq.}{Eqs.}
\crefname{figure}{Figure}{Figures}
\crefname{table}{Table}{Tables}
\crefname{algorithm}{Algorithm}{Algorithms}

\usepackage{url}
\usepackage{epsfig}
\usepackage{amssymb}
\usepackage{etoolbox}
\usepackage{multirow}
\usepackage{listings}
\usepackage{enumitem}
\usepackage{booktabs}
\usepackage[most]{tcolorbox}
\usepackage{bm}        
\usepackage{dsfont}    
\def\eg{\emph{e.g.,}}

\def\etc{\emph{etc.}}
\def\ie{\emph{ie.,}}

\usepackage{algorithm,algorithmicx,algpseudocode}
\usepackage[table]{xcolor}
\RequirePackage{xcolor} 

\title{\Ourmodel: Explicit Layout Planner for Easier LLM-Driven \\ 3D Room Generation }
\newcommand{\corrauthor}{\textsuperscript{\rm $\dagger$}}

\author{
    Wenzhuo Sun\textsuperscript{\rm 1}\equalcontrib,
    Mingjian Liang\textsuperscript{\rm 1}\equalcontrib,
    Wenxuan Song\textsuperscript{\rm 2},
    Xuelian Cheng\textsuperscript{\rm 1}\corrauthor,
    Zongyuan Ge\textsuperscript{\rm 1}
}
\affiliations{
    \textsuperscript{\rm 1}Monash University, Melbourne, Australia\\

    \textsuperscript{\rm 2}The Hong Kong University of Science and Technology(GZ), Guangzhou, China
}

\usepackage{bibentry}

\begin{document}
\maketitle
\author{}

\input{sec/0_abstract}    
\input{sec/1_intro}

\input{sec/2_related}
\input{sec/preliminary}
\input{sec/3_method}
\input{sec/4_experiments}
\input{sec/5_conclusion}

\noindent
\begin{minipage}{\textwidth}
    \centering
    {\LARGE\bfseries Appendix Material \par}
\end{minipage}
\vspace{0.5em}
\input{sec/6_appendix}
\newpage
{
    \small
    \bibliography{main}
}

\end{document}

%% file: sec/0_abstract.tex
\begin{abstract}


In this paper, we propose \textit{\Ourmodel}, the first fully automatic 3D room generation framework for painlessly creating realistic indoor scenes with only short text as input. Without any manual layout design or panoramic image guidance, our framework can generate explicit layout criteria for rational spatial placement. We begin by introducing a hierarchical structure of language-driven agent planners that can automatically parse short and ambiguous prompts into detailed scene descriptions. These descriptions include raw spatial and semantic attributes for each object and the background, which are then used to initialize 3D point clouds. To position objects within bounded environments, we implement two arrangement constraints that iteratively optimize spatial arrangements, ensuring a collision-free and accessible layout solution.
In the final rendering stage, we propose a novel AnyReach  Sampling strategy for camera trajectory, along with the  Interval Timestep Flow Sampling (ITFS) strategy, to efficiently optimize the coarse 3D Gaussian scene representation. These approaches help reduce the total generation time to under 30 minutes. 
Extensive experiments demonstrate that our method can produce geometrically rational 3D indoor scenes, surpassing prior approaches in both rendering speed and visual quality while preserving editability. The code will be available soon.
  
\end{abstract}


%% file: sec/1_intro.tex
\section{Introduction}

Text-to-3D scene generation aims to convert textual descriptions into 3D environments. Designed to lower costs and reduce the required expertise, this paradigm is conducive to various downstream applications, such as embodied AI~\cite{chen2024urdformer,embodiedscan}, gaming~\cite{wonderworld}, filmmaking, and augmented / virtual reality~\cite{Real-time3D}. Benefitting from the remarkable success of generation models~\cite{diffusion, ddim} and implicit neural representation~\cite{nerf,3DGS}, numerous methods have emerged to address the challenges of text-to-3D object generation, achieving significant milestones. However, directly transitioning from generating 3D assets to creating complete scenes is not a trivial task. This is partly due to the fact that layout information requires additional context and human effort. Based on the extracted layout sources, existing 3D scene generation methods can be broadly classified into two categories: \textbf{I.}~\emph{Visual-guided}, and \textbf{II.}~\emph{Rule-based} methods. 

\begin{figure*}
  \centering
  \includegraphics[width=1.0\textwidth]{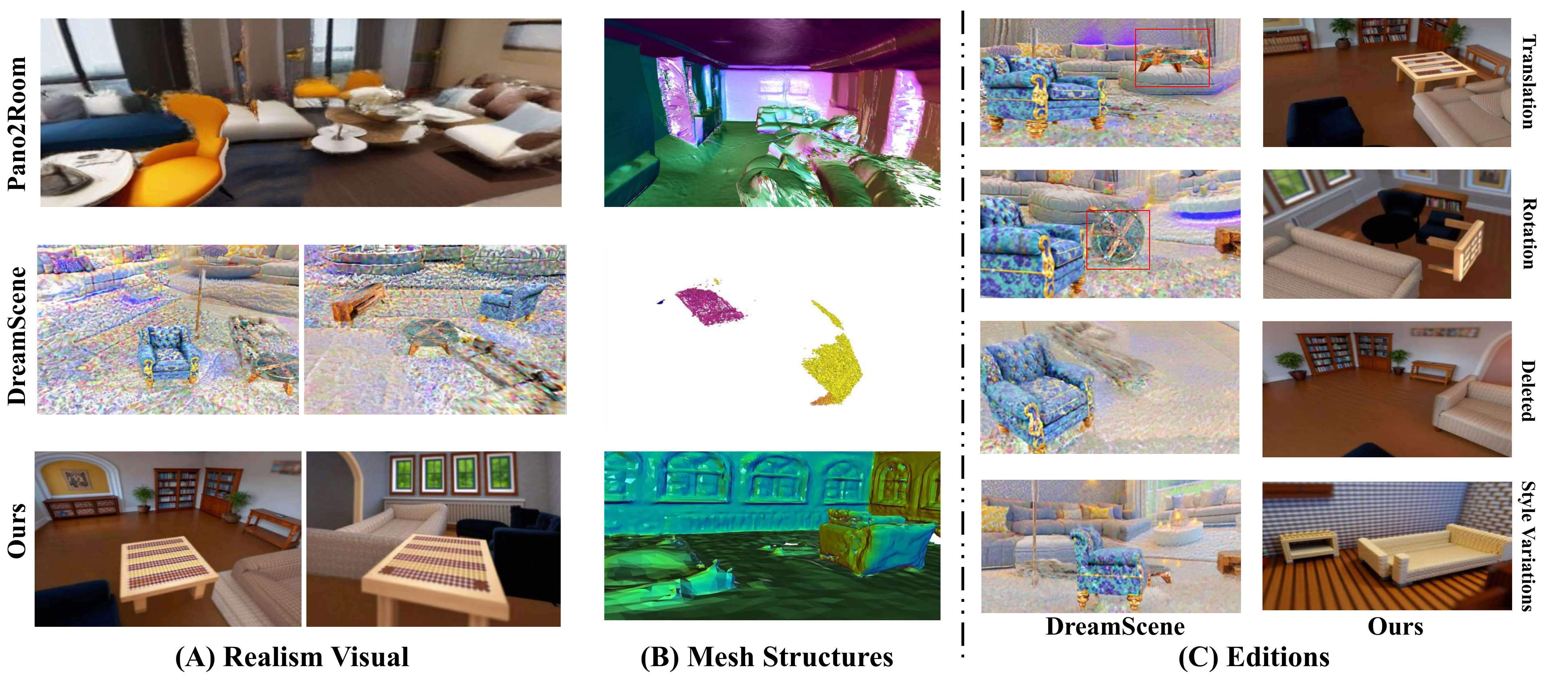}
  \caption{Compared to previous methods,~\ie visual-guided method Pano2Room~\cite{pu2024pano2room} and rule-based method DreamScene~\cite{dreamscene}, our approach effectively generates indoor scenes characterized by (a) more realism, (b) smoother mesh structures, and (c) support for a diverse array of editions, including operations such as rotation, translation, importing/deleting 3D assets, and style variations. }
  \label{fig:teaser}
\end{figure*}

Visual-guided methods rely on extracting implicit spatial information from visual representations, \eg~multi-view RGB or panorama images. Methods in this category typically adopt a two-stage generation framework, where image generation serves as an intermediate process, followed by an implicit neural representation network that renders a 3D scene from previously generated images. Some methods~\cite{director3d, scene4u} treat objects and backgrounds as an indivisible whole as input for the implicit rendering network. As a result, these continuous representations produced by these methods lack object-level decoupling, which significantly restricts practical use. These kinds of approaches can also lead to foggy or distorted geometry, as shown in~\cref{fig:teaser} (b). 
Instead of being directly constrained, the layout configuration is learned from a text-to-image generation model, which is optimized by aligning the generated images with the input text using a similarity score. Thus, the optimization of the scene configuration is constrained by the capabilities of the image generation model, making it difficult to directly control the layout through textual descriptions. 

Rule-based methods, on the other hand, explicitly utilize predefined rules based on physical relationships and constraints to create layout information of 3D environments.
Early methods~\cite{dreamscene,GALA3D,setthescene,Text2Room} depend on manually crafted prompt templates or human-designed layouts for initialization, which helps generate physically plausible scenes. This procedure relies on specialized expert knowledge and typically involves a tedious trial-and-error process to refine predefined spatial attributes of objects,~\eg~ position, size, and rotation~\etc. This human-in-the-loop paradigm increases the demand for human labor while also undermining the end-to-end training process.

On a separate line of research and to reduce human effort in layout design, data-driven approaches treat scene synthesis as a 3D asset arrangement task within a bounded environment. These methods add objects sequentially using an order prior with a forest structure ~\cite{sun2024forest2seq} or simultaneously through conditional score evaluation ~\cite{maillard2024debara}. Recently, Large Language Models (LLMs) have been introduced to automate the creation of 3D scenes~\cite{holodeck2023, yang2024llplace, Sun_2025_CVPR}. LLMs can directly translate unstructured language into structured spatial representations, which reduces dependence on rigid templates. However, their reliance on existing 3D assets limits the versatility and flexibility required to generate diverse and innovative scenes.

In this paper, we present \textbf{\Ourmodel}, which combines the advances of visual-guided and rule-based models. Unlike previous approaches, our method enables automatic generation of differentiable 3D scenes and layout constraints given only short textual prompts. Our framework adopts a `Reasoning-Grounding, Arrangement, and Optimization' pipeline to break down the complex 3D scene generation task into controllable and scalable subtasks. In the reasoning and grounding stage, our hierarchical LLM-driven agent planners can parse short user prompts into executable long-term descriptions. Specifically, our high-level LLM agent produces textual descriptions that select the initial target scene and objects that satisfy the contextual semantics. The low-level LLM agent generates fine-grained grounding descriptions that infer object and background scales, along with layout rules that match real-world dimensions. These explicit physical constraints also ensure that the generated 3D scenes meet the interaction requirements.

After obtaining the initial scene description from the LLM agents, we utilize a text-to-3D object generator~\cite{shap-e} to create 3D assets represented as point clouds. To assign initial object assets into the scene, we then employ interactive optimization to update the layout design by verifying whether the generated layout meets our constraints. To improve the rationality of the indoor scene, we first apply collision constraints to detect overlaps between objects. Additionally, we introduce a reachability constraint to assess path viability between entry points and target assets, supporting dynamic asset placement through closed-loop optimization. The generated scenes remain fully editable at both the object and global levels, allowing for reposition, insertion, deletion, and style transfer, as shown in~\cref{fig:teaser} (c). 

For photorealistic rendering, we introduce a module that integrates Interval Timestep Flow Sampling (ITFS) with reflection flow matching and hierarchical multi-timestep sampling. By leveraging a 2D diffusion prior and detailed scene descriptions from LLM agents, this module generates continuous 3D representations that exhibit a realistic appearance and rich semantics. We also propose the AnyReach camera trajectory sampling strategy that can optimize both object-centric and global views within a single rendering pass. Consequently, our pipeline produces high-fidelity scenes in a single optimization stage, reducing the overall runtime by nearly 2× compared to prior work.

%% file: sec/2_related.tex
\section{Related Work}
\label{rw}

\subsection{Text-to-3D Object Generation}
Existing 3D scene generation methods can be broadly classified into two categories: \textbf{I.}~\emph{Visual-Guided}, and \textbf{II.}~\emph{Direct regression} methods. 
Visual-Guided methods~\cite{prolificdreamer,luciddreamer,Trellis} rely on pretrained 2D diffusion models, which can lead to error accumulation during the subsequent image-to-3D conversion process. While MagicTailor~\cite{magictailor} addresses the issue of semantic pollution in the text-to-image stage, the resulting continuous representations still struggle with object-level decoupling. Direct regression methods~\cite{point-e,shap-e} focus on generating 3D assets directly from input text. Although 3D object generation has made significant advancements, directly applying a similar philosophy to solve 3D scene generation is still very challenging. Due to the lack of high-quality text-to-3D scene datasets, it hinders the ability to implement a feed-forward network directly. Furthermore, incorporating layout information adds an additional layer of complexity to the scene synthesis process.

\subsection{LLM driven 3D Scene Generation}
As discussed in \textsection Introduction, rule-based methods can utilize explicit layout information to generate accurate and diverse 3D layouts. However, designing an effortless mechanism for layout arrangement remains an open question in 3D scene synthesis. With the reasoning capabilities of LLMs, works such as LayoutGPT~\cite{layoutgpt} and SceneTeller~\cite{sceneteller} can generate visual layouts and place existing 3D assets for scene synthesis. However, because of the single-step planning strategy, their output often exhibits physical implausibilities,~\eg~floating objects or intersecting geometries. Additionally, they tend to perform suboptimally in complex scenes with numerous objects, mainly due to the absence of spatial constraints. 

On the other hand, LLM-driven frameworks aim to reduce human effort in the 3D layout arrangement task. To improve layout rationality, recent reinforcement learning-based methods utilize constraint solvers to enhance physical validity. Holodeck~\cite{holodeck2023} and its extensions~\cite{yang2024llplace,Sun_2025_CVPR} translate relational phrases into geometric constraints, which are optimized using gradient-based methods to eliminate collisions.  Diffuscene~\cite{diffuscene} and Physcene~\cite{physcene} incorporate physics engines to simulate object dynamics, iteratively adjusting parameters derived from LLMs to ensure stability. Despite advances in realism, their reliance on existing 3D assets limits the versatility and flexibility needed for generating diverse and innovative scenes. Beyond these layout arrangement approaches, our model provides a fully automated pipeline for generating complete and controllable 3D scenes, including individual decoupling objects.


%% file: sec/preliminary.tex
\section{Preliminary}

\paragraph{3D Layout Arrangement}

Taking layout descriptions as input, this problem aims to arrange unordered 3D assets from existing datasets within a 3D environment. Formally, given a layout criterion \( \varphi_{\text{layout}} \), a space defined by four walls oriented along the cardinal directions $\{w_1, \dots, w_4\}$, and a set of \( N \) 3D meshes $\{m_1, \dots, m_N\}$, the objective is to create a 3D scene that reflects the most accurate spacial relationships of the provided layout configuration. Previous methods~\cite{holodeck2023,feng2023layoutgpt,Sun_2025_CVPR} treat indoor layout as 3D reasoning tasks, arranging objects in space according to open-ended language instructions. They assume that the input 3D objects are upright and an off-the-shelf vision-language model (VLM),~\eg~GPT-4~\cite{gpt4}, to determine the front-facing orientations of the objects. As one of the representative methods, LayoutVLM~\cite{Sun_2025_CVPR} annotates each object with a concise textual description \( s_i \), and the dimensions of its axis-aligned bounding box, after rotating to face the \( +x \), are represented as \( b_i \in \mathbb{R}^3 \). The desired output of the layout generation process is the pose of each object \(p_i = (x_i, y_i, z_i, \theta_i)\), which includes 3D position and its rotation about the \( z \)-axis.

\paragraph{Neural Representation} 
Unlike explicit representations such as meshes or point clouds, implicit neural representation methods train a 3D model through differentiable rendering. There are mainly two types of neural representations: Neural Radiance Fields~\cite{nerf} and 3D Gaussian Splatting(3DGS)~\cite{3DGS}. 3DGS represents the scene as a set of 3D Gaussians $\{\mathcal{G}_{i}\}_{i=1}^{M}$, each parametrized with center position $\mathbf{\mu}_i \in \mathbb{R}^{3}$, covariance $\mathbf{\Sigma}_i \in \mathbb{R}^{3\times 3}$, color $c_i \in \mathbb{R}^{3}$, and opacity $\alpha_i \in \mathbb{R}^{1}$. The 3D Gaussians can be queried by $\mathcal{G}(x)~= e^{-\frac{1}{2}(x)^{T}\Sigma^{-1}(x)}$, where $x$ represents the distance between $\mathbf{\mu}$ and the query point. The 3D Gaussians are optimized through differentiable rasterization for projection rendering, comparing the resulting image to the training views in the captured dataset using image loss metrics. In the text-to-3D task, there is typically no ground truth (GT) image available. Existing works usually employ a diffusion model to generate pseudo-GT images for distilling the 3D representation.

\paragraph{Diffusion Models}
Diffusion models~\cite{diffusion,ddim} generate data by iteratively denoising a sample from pure noise.  During training, a noisy version of a data sample $\mathbf{x} \sim p_{\text{data}}$ is generated as $\mathbf{x}_{t} = \sqrt{\bar{\alpha}_{t}}\,\mathbf{x} + \sqrt{1 - \bar{\alpha}_{t}}\bm{\epsilon}$, where $\bm{\epsilon} \sim \mathcal{N}(\mathbf{0}, \mathbf{I})$ is standard Gaussian noise, and $\bar{\alpha}_{t}$ controls noise level. The discrete diffusion timestep $t$ is sampled from a uniform distribution $p_t \sim \mathcal{U}(0, t_{max})$. The denoising network $\theta$ predicts the added noise $\bm{\epsilon}_{\theta}$ and is optimized with the score matching objective:
$\mathcal{L}_t=\mathbb{E}_{\mathbf{x}\sim p_{\text{data}},\, t \sim p_t,\, \bm{\epsilon}\sim \mathcal{N}(\mathbf{0}, \mathbf{I})}\left[\left\lVert \bm{\epsilon} - \bm{\epsilon}_{\theta}(\mathbf{x}_{t}, t)\right\rVert_2^2\right]$.

DreamFusion~\cite{dreamfusion} distills 3D representations from a 2D text-to-image diffusion model through Score Distillation Sampling (SDS). Let $\theta$ denote the parameters of a differentiable 3D representation, and let $g$ represent a rendering function. The rendered image produced for a given camera pose $c$ can be expressed as $x_0 = g(\theta, c)$, where $x_0$ is the clean (noise-free) rendering.  A noisy view at timestep $t$ is obtained as: $ x_t = \alpha_t\,x_0 + \sigma_t\,\boldsymbol{\epsilon}.$
Then SDS distills $\theta$ through a 2D diffusion model $\phi$ with frozen parameters as follows:
\begin{equation*}
\label{eqn:sds}
    \begin{aligned}
    \nabla_{\theta} \mathcal{L}_\text{SDS}(\theta) = 
    \mathbb{E}_{t, \boldsymbol{\epsilon}, c} 
    \Bigg[ 
        & w(t) \left( \boldsymbol{\epsilon}_\phi(\mathbf{x}_t; y, t) 
        - \boldsymbol{\epsilon} \right)
        \times \frac{\partial g(\mathbf{\theta}, c)}{\partial \theta}
    \Bigg], 
    \end{aligned}
\end{equation*}
where $w(t)$ is a timestep-dependent weight and $\boldsymbol{\epsilon}_{\phi}$ is the noise estimator of the frozen diffusion model~$\phi$ conditioned on the text prompt $y$.


%% file: sec/3_method.tex
\section{Methodology}
\label{method}

\begin{figure}[t!]
  \centering
  \includegraphics[width=0.42\textwidth]{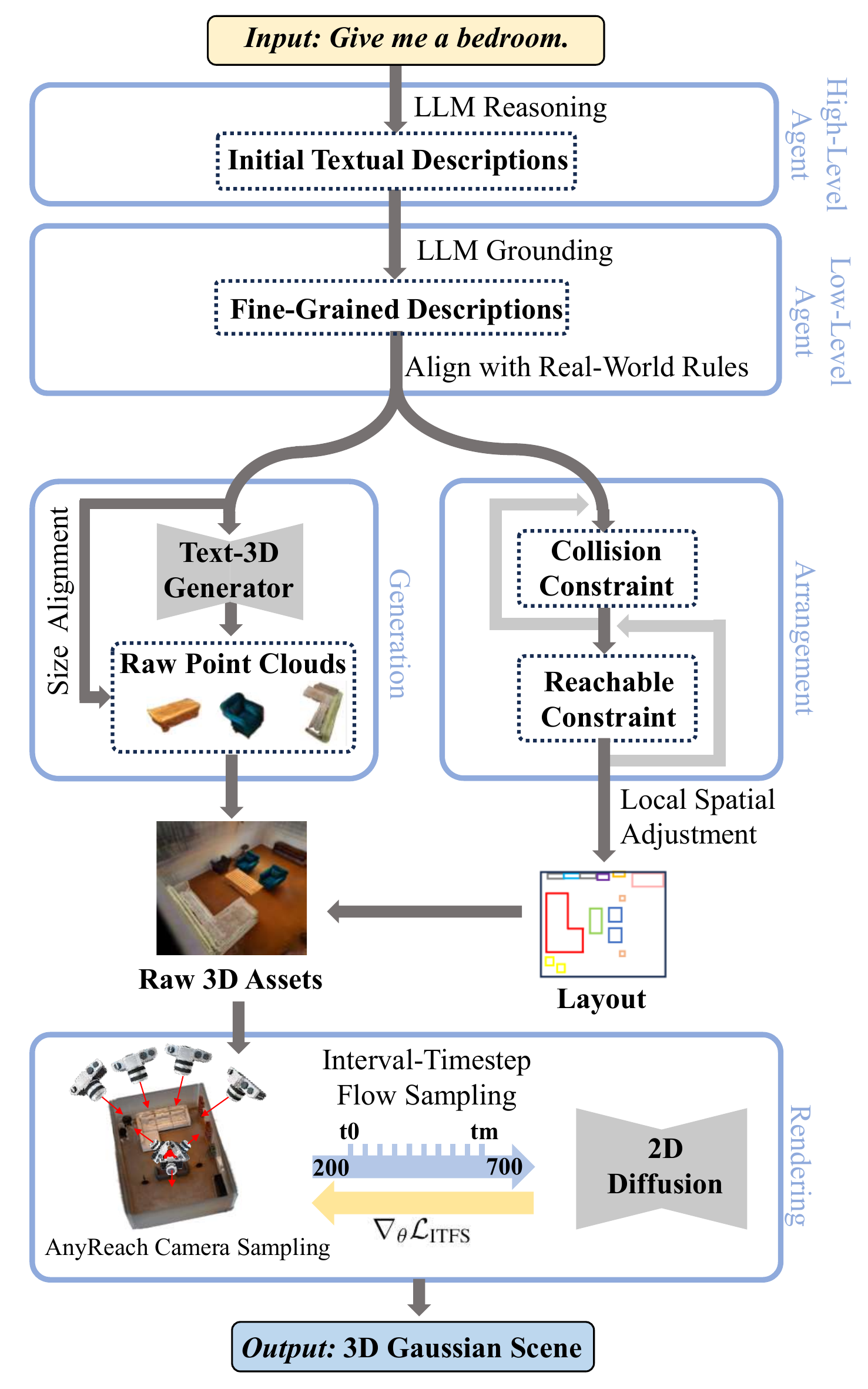}
  \caption{~\Ourmodel~follows a `Reasoning-Grounding, Arrangement, and Optimization' pipeline, decomposing the complex 3D scene generation task into scalable subtasks. }
  \label{fig:pipeline}
\end{figure}

\Ourmodel~is a fully automated pipeline designed to generate complete, complex, and controllable 3D indoor scenes from concise user prompts. As shown in~\cref{fig:pipeline}, our framework comprises three main components. We begin by introducing a hierarchical LLM-driven planner that reasons and grounds the target room type based on the user prompt. It produces detailed textual descriptions ${y_i}{(i=1,...,N)}$ for each asset and a scene layout configuration $y_\text{scene}$. Next, we create the initial 3D point cloud for each asset using a text-to-3D object generator~\cite{shap-e} with $y_i$ as input. Meanwhile, a layout planner computes the optimal sizes and poses for the assets, generating a complete scene map that includes object coordinates, orientations, physical properties, and style descriptions. The assets are then positioned within the coordinate system of the room, according to pose specifications. Finally, we employ an Interval Timestep Flow Sampling with a novel camera sampling strategy to perform single-stage scene optimization, yielding results with realistic visuals and high 3D consistency.

\begin{algorithm}[t]
\caption{~\Ourmodel~Framework}
\label{alg:framework}
\begin{algorithmic}[1]
\State \textbf{Input}: short‑prompt description $\{\mathbf{I}\}$
\State \textbf{Output}: 3D room scene $\mathbf{Scene}^{\star}$

\State \textcolor{gray}{// High‑Level Reasoning}
\State $(\mathcal{O},\mathcal{S}) \gets \textsc{LLMParse}(\mathbf{I})$

\State \textcolor{gray}{// Low‑Level Grounding}
\State $\mathcal{P} \gets \textsc{InitPointCloud}(\mathcal{O},\mathcal{S})$
\State $(\hat{\mathcal{O}},\hat{\mathcal{S}}) \gets
       \textsc{AlignSizes}(\mathcal{O},\mathcal{S},\mathcal{P})$

\State \textcolor{gray}{// Layout Arrangement I: Collision Constraint}
\State $\varphi_{\text{layout}} \gets
       \textsc{Arrange}(\hat{\mathcal{O}},\hat{\mathcal{S}},\mathcal{P})$
\For{$t_1 = 1$ to $\text{maxTimeIter}$}
    \If{\textbf{not}\; $\textsc{CollisionFree}(\varphi_{\text{layout}})$}
        \State $\varphi_{\text{layout}} \gets \textsc{FixLayout}(\varphi_{\text{layout}},\textit{collision})$
    \Else
        \State \textbf{break} \Comment{collision‑free layout $\hat{\varphi}_{\text{layout}}$ found}
    \EndIf
\EndFor

\State \textcolor{gray}{// Layout Arrangement II: Reachability Constraint}
\For{$t_2 = 1$ to $\text{maxTimeIter}$}
    \If{\textbf{not}\; $\textsc{Reachable}(\hat{\varphi}_{\text{layout}})$}
        \State $\hat{\varphi}_{\text{layout}} \gets \textsc{FixLayout}(\hat{\varphi}_{\text{layout}},\textit{reachable})$
    \Else
        \State \textbf{break} \Comment{reachable layout $\varphi_{\text{layout}}^{\star}$ found}
    \EndIf
\EndFor

\State \textcolor{gray}{// Differentiable Scene Rendering}
\State $\mathcal{T} \gets \textsc{AnyReach}(\varphi_{\text{layout}}^{\star})$
\State $\mathbf{Scene}^{\star} \gets \textsc{ITFS}(\varphi_{\text{layout}}^{\star},\mathcal{T})$
\end{algorithmic}
\end{algorithm}

\subsection{Hierarchical LLM Agents Planning}
Given a concise user prompt \( T \), such as \textit{``A bedroom''}, our high-level planner adopts the reasoning capabilities of an LLM agent to semantically expand \( T \) and identify relevant object categories for the scene population. Select the appropriate types of objects that align with the scene style specified by the input prompt. The LLM agent also generates an indexed set $(O_i, S_i )$ for the room. Subsequently, the low-level LLM agent planner is used to enhance these assets with stylistic, textural, and material attributes. 

For spatial information, we first adopt Shap-E~\cite{shap-e}  to generate the corresponding point cloud representation  $\mathbf{p}_i=(x_i, y_i, z_i, c_i)$ for each object \(O_i\) according to the index $i$. We then predict the central position for each object by estimating candidate center coordinates $(x^{c}_i, y^{c}_i, z^{c}_i)$ in the scene. Each object $O_i$ is associated with a semantic label. To ensure semantic and physical plausibility, the physical agent aligns the real-world scale of each object and rescales it if inconsistencies are detected. The updated object parameters and spatial context $(\hat{O}_i, \hat{S}_i )$ are stored in \textit{.json} files for downstream tasks. Further details can be found in the Supplementary Material (Supp).

\subsection{Layout Arrangement Planning}
Taking the detail layout criterion $\varphi_{\text{layout}}=(\hat{O}_i, \hat{S}_i, \mathbf{p}_i)$ as input, we treat the indoor scene generation task as if in a closed space, following the same strategy used in a previous 3D layout arrangement method~\cite{holodeck2023}. The process begins with the construction of structural elements such as floors, walls, doors, and windows, followed by the arrangement of floor-mounted furniture and the placement of wall-mounted objects.
Next, we employ interactive optimization to update the layout design by verifying whether the generated layout satisfies our two constraints, which are specified as follows.

\paragraph{Collision Constraint} 
Given the inferred spatial constraints $\varphi_{\text{layout}}$ by LLM agents, we first adopt the depth-first-search algorithm to find a valid placement for object candidates inspired by Holodeck~\cite{holodeck2023}. We sequentially place objects according to the inferred symbolic rules, such as positioning objects close to walls, aligning them with other objects(~\eg~\textit{nightstands} beside a \textit{bed}), or oriented in specific directions. To enhance layout quality, we further decouple wall-mounted and ground-level object types while ensuring they both follow the same symbolic reasoning process.
Specifically, we define a collision-aware layout reward $\mathcal{R}_{\text{coll}}$ to ensure physical feasibility.

\begin{equation*}
\begin{aligned}
\mathcal{R}_{\text{coll}} =
- (\sum_{i \ne j} \text{IoU}_{3D}(b_i, b_j) + 
\sum_{i=1}^N \sum_{k=1}^{W} \text{IoU}_{3D}(b_i, \text{b}_k^{wall})),
\end{aligned}
\label{eq:coll_guide}
\end{equation*}
where $b_i$ and $b_j$ represent the bounding boxes of objects, and $b_k^{wall}$ denotes the bounding boxes of wall structures.
The $\text{IoU}_{3D}$ is a metric that measures the discrepancy between two candidate 3D bounding boxes. Note that a layout is valid only when $\mathcal{R}_{\text{coll}} = 0$. In practice, we set a time limit of 30 seconds for the $t_1$ iterating process; after this period, any objects that are still colliding will be removed.

\paragraph{Reachability Constraint} 
We ensure navigability by requiring all objects to be accessible to a virtual agent represented by the bounding box $b^{agent}$. Different from the reachability constraints in Physcene~\cite{physcene}, we rasterize the 3D Scene Map into a 2D traversable map and employ the A* search algorithm, initiating from a fixed starting point (~\eg~a doorway) to each object location. An object is considered reachable only if a valid path exists. The reachability reward $\mathcal{R}_{\text{reach}}$ is defined as follows:

\begin{equation*}
\mathcal{R}_{\text{reach}} = - \sum_{i=1}^{W} \text{IoU}_{3D}(b_i, b^{agent}).
\label{eq:agent_guide}
\end{equation*}
If an object is deemed unreachable, we undertake local spatial adjustments within a restricted search grid. Again, if no viable solution is identified in 30 seconds, the object will be removed from the layout. This integrated reasoning process ensures both physical plausibility and functional accessibility of the final layout $\varphi_{\text{layout}}^{\star}$.

\subsection{Differentiable Scene Optimization}
\paragraph{AnyReach Camera Trajectory Sampling}
Proir work AnyHome~\cite{anyhome} hypothesizes that the camera orients toward the room's center, generating egocentric trajectories that spiral around the scene. Meanwhile, they randomly sample the camera view to capture local details and refine object-level layouts. Unlike AnyHome, our AnyReach sampling strategy supports a more accurate ``Zoom-in and Zoom-out" approach to swiftly capture global and local views. 

Specifically, we utilize a spiral camera path across the upper hemisphere surrounding each object, with each camera directed toward the object's center. This spiral sampling aims to provide a global observation of the background and objects within the scene. When the ``Zoom-in" movement occurs, our camera employs the A* algorithm~\cite{Astar} to navigate to the object's location, ensuring the shortest path from the global view to the local object-level view. Taking advantage of A*, our approach conserves camera poses and minimizes the time spent observing irrelevant background such as \textit{wall} along the way. Once the camera reaches the object, we implement a ``Zoom-out" movement directly from the current object location to the original global path. This repeated jumping strategy prevents prolonged focus on a single object, ensuring comprehensive view supervision and reducing the object-level geometric blur from distant viewing. More visualization and implementation details are provided in the Supp.




\paragraph{Interval Timestep Flow Sampling}  
In the rendering stage of~\cref{fig:pipeline}, to enhance the quality of raw 3D scene representations, we propose an Interval-Timestep Flow Sampling (ITFS) strategy that utilizes a range of different timesteps for generating 2D diffusion views. Our approach integrates ITFS with reflection flow matching and multi-timestep sampling, gradually optimizing scene quality from coarse to fine with enhancing semantic details (visualized in~\cref{fig:experiment_layout}).

Conditional Flow Matching (CFM)~\cite{lipman2023flow} has proven to be a robust framework for training continuous normalizing flows, primarily due to its efficiency in computation and overall effectiveness. To obtain higher-fidelity priors, we replace $\phi$ with Stable Diffusion 3.5~\cite{scaling}, which utilizes a rectified-flow formulation to model both the forward and reverse processes through a time-dependent vector field. The forward interpolation is defined as $x(t) = (1-t)\,x_0 + t\,\boldsymbol{\epsilon},~\boldsymbol{\epsilon}\sim\mathcal{N}(\mathbf{0},\mathbf{I})$, and the reverse field $v(x,t)=\tfrac{\partial x}{\partial t}$ is approximated by a neural network $v_{\phi}$. The training process adheres to the CFM objective:
\begin{equation*}
\mathcal{L}_\text{CFM}(\theta) = 
\mathbb{E}_{t, \boldsymbol{\epsilon}} \left[
    w(t) \left\|
        v_\phi(x_t; t) - (\boldsymbol{\epsilon} - x_0)
    \right\|_2^2
\right]
\end{equation*}

Building on the CFM objective function and SDS, we introduce a new \textbf{Flow Distillation Sampling (FDS)} strategy. The objective function $\nabla_{\theta}\mathcal{L}_\text{FDS}$ is defined as follows: 
\begin{equation*}
\label{eqn:x0-hatx0}
    \begin{aligned}
          \mathbb{E}_{t,\boldsymbol{\epsilon}}\Big[
          w(t)\bigl( v_\phi(x_t; y, t) -
          (\boldsymbol{\epsilon} - x_0) \bigr) \times \frac{\partial g(\theta, c)}{\partial \theta}
    \Big]
    \end{aligned}
\end{equation*}
Empirically, timesteps $t\in[200, 300]$ emphasize geometric features, while larger timesteps $t >500$ favor semantic alignment, a pattern also noted by DreamScene~\cite{dreamscene}. Since FDS samples in a single timestep, it cannot take advantage of both regimes. To address this limitation, we introduce ITFS which generates intermediate steps from small timestep intervals $t_0$ to large timestep intervals $t_m$. This is achieved during the optimization of the objective function  $\nabla_{\theta}\mathcal{L}_{\text{ITFS}}$:
 
\begin{equation*}
\label{eq:itfs}
    \begin{aligned}
    \mathbb{E}_{t,\boldsymbol{\epsilon}}\!\Bigg[
           \sum_{i=1}^{m} w(t_i)\,
           \bigl( v_\phi(x_t; y, t_i) - (\boldsymbol{\epsilon} - 
           x_0) \bigr) \times \frac{\partial g(\theta, c)}{\partial \theta}
         \Bigg] ,
    \end{aligned}
\end{equation*}

where $m$ denotes the value of the sampling numbers. This multi-interval strategy allows early steps to refine geometric features while later steps consolidate semantic information. As a result, it produces 3D scenes that are both structurally accurate and visually coherent.


%% file: sec/4_experiments.tex
\begin{table*}[t!]  
\centering
\setlength{\tabcolsep}{1.4mm}
\small 
\begin{tabular}{c|c|c c c c c c c}
        \toprule
        Type &  Method & \shortstack{Layout \\ Planning}  & \shortstack{End-to-End \\ Generation} & \shortstack{Room \\ Scene} & \shortstack{Edition} & \shortstack{Realistic \\ (object)} & \shortstack{Realistic \\ (scene)} & \shortstack{Surface \\ Reconstruction} \\
        \midrule
        
        \multirow{4}{*}{Visual-Guided} 
        & Text2Room~\cite{Text2Room} & - & - & - & - & \ding{51} & \ding{51} & \ding{51} \\
        & SceneWiz3D~\cite{Zhang_2024_CVPR} & - & - & - & \ding{51} & -  & - & \ding{51} \\
        & Pano2Room~\cite{pu2024pano2room} & - & - & - & \ding{51} & -  & - & \ding{51} \\
        & Director3d~\cite{director3d} & - & - & - & \ding{51} & -  & - & \ding{51} \\
        & Scene4U~\cite{scene4u} & - & - & \ding{51} & - & -  & - & \ding{51} \\
        \midrule

        \multirow{6}{*}{Rule-based} 
        & Set-the-Scene~\cite{setthescene} & \ding{51} & - & - & - & \ding{51} & \ding{51} & \ding{51} \\
        & GALA3D~\cite{GALA3D} & \ding{51} & \ding{51} & - & \ding{51} & \ding{51} & -& - \\
        & AnyHome~\cite{anyhome} & \ding{51} & - & \ding{51} & \ding{51} & \ding{51} & -  & \ding{51} \\
        & GraphDreamer~\cite{graphdreamer} & \ding{51} & \ding{51} & - & \ding{51} & \ding{51} & -  & - \\
        & DreamScene~\cite{dreamscene} & \ding{51} & \ding{51} & - & \ding{51} & \ding{51} & \ding{51}& - \\
        
        \midrule
        \rowcolor{lightgray!30}
        & Ours & \ding{51} & \ding{51} & \ding{51} & \ding{51} & \ding{51} & \ding{51} & \ding{51} \\
        \bottomrule
    \end{tabular}
    \caption{Comparison of 3D indoor scene synthesis methods, categorized into rule-based and visual-guided approaches. \ding{51} means capability supported; `-' means without that capability. Our method demonstrates comprehensive capabilities across all metrics.}
\label{tab:dataset-table}
\end{table*}

\begin{figure*}[t!]
  \centering
  \includegraphics[width=1.0\textwidth]{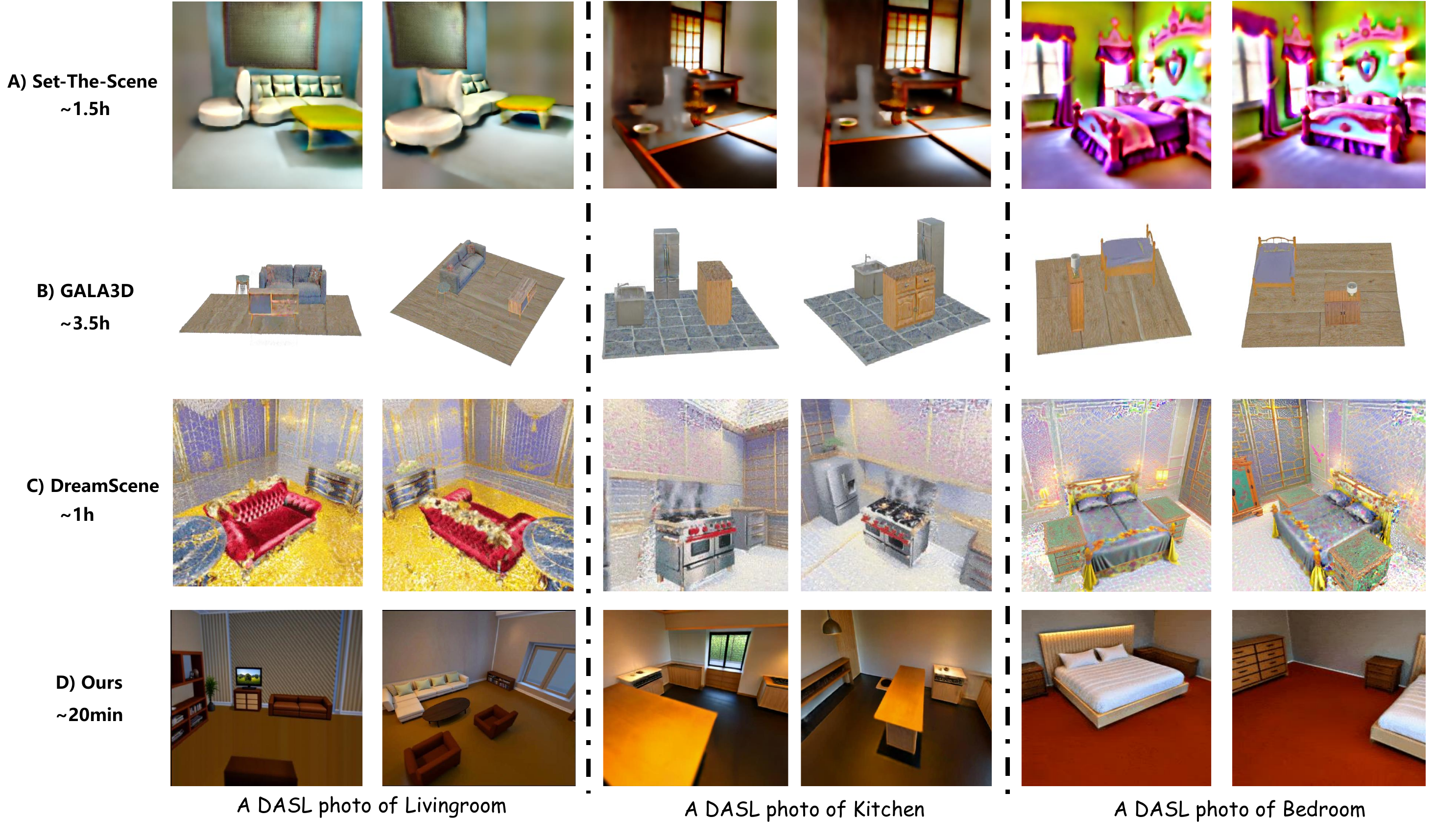}
  \caption{We compare the interactive scene 3D generation with Set-the-Scene~\cite{setthescene}, GALA3D~\cite{GALA3D} and DreamScene~\cite{dreamscene}.}
  \label{fig:experiments}
\end{figure*}

\section{Experiments}
\label{experiments}
\paragraph{Implementation Details}
We utilize GPT-4~\cite{gpt4} as our LLM agents for both reasoning and grounding modules. Shap-E~\cite{shap-e} generated point clouds serve as the initial representation of objects, while Stable Diffusion 3.5 Medium~\cite{StabilityAI23} provides guidance for rendering 3D scene models. We set the rendering iterations to 1500 and the resolution to 512 $\times$ 512, and ITFS sampling interval value $m=3$, drawing $t_0\!\sim\![200,400]$, $t_1\!\sim\![400,600]$ and $t_2\!\sim\![600,800]$.  We tested~\Ourmodel~and all baselines on the same NVIDIA 4090 GPU for fair comparison.

\subsection{Qualitative Results}
\cref{fig:experiments} compares our model with other state-of-the-art methods. We observe that GALA3D~\cite{GALA3D} lacks adequate asset diversity, while Set-the-Scene~\cite{setthescene} shows geometric discontinuities. Additionally, DreamScene~\cite{dreamscene} suffers from particle opacity artifacts and coarse surface geometry, all of which compromise visual fidelity. In contrast, our model generates physically plausible scene synthesis. The decoupled guidance and layout arrangement approaches ensure physical interaction constraints without sacrificing scene diversity. As a result, our model can produce high-quality 3D scenes with object-level decoupling, surpassing prior differentiable approaches in rendering speed, visual quality, and geometric consistency, all while maintaining editability.

\begin{figure}[t]
  \centering
  \includegraphics[width=0.48\textwidth, height=0.40\textwidth]{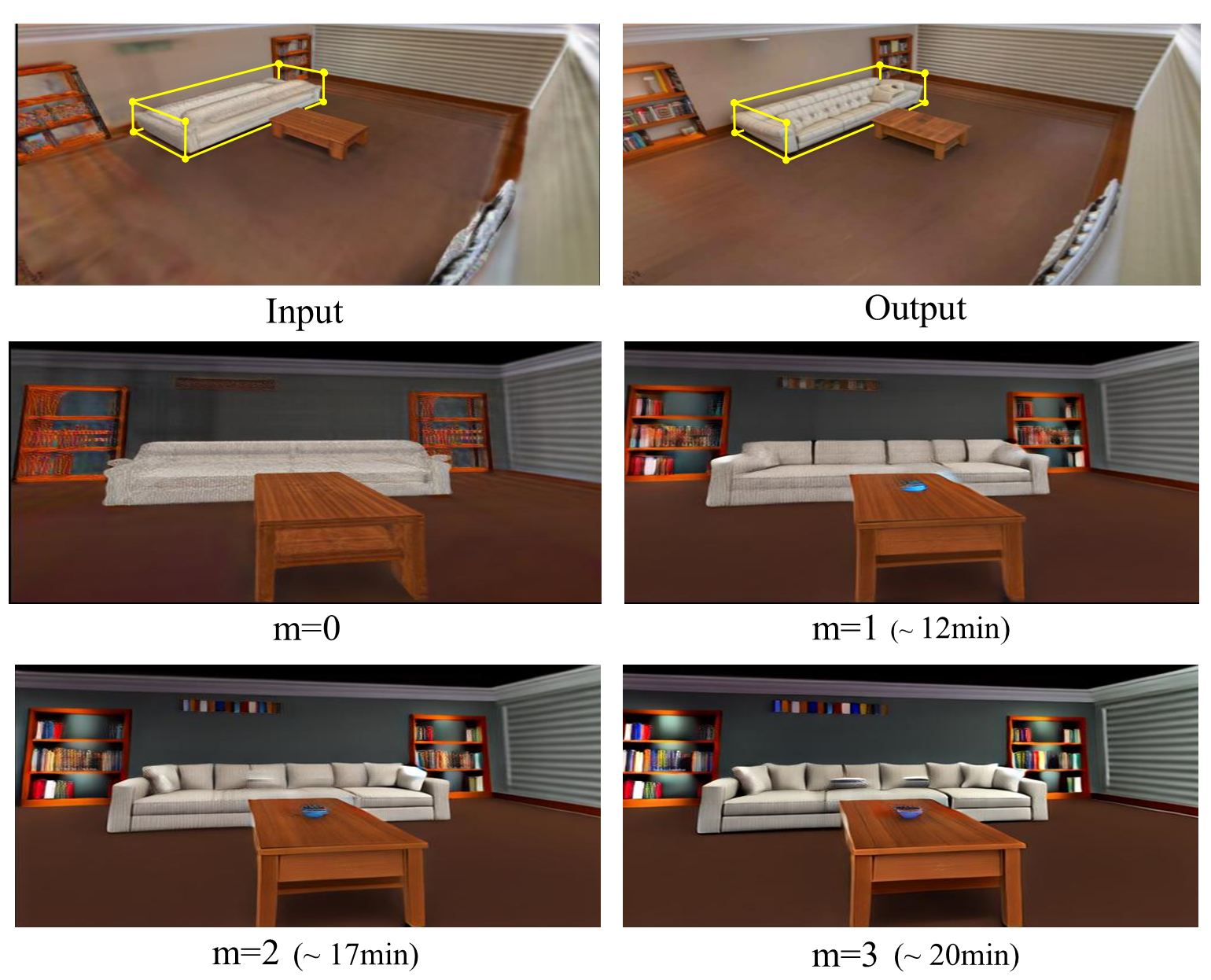}
  \caption{Effectiveness of ITFS. The coach labels in yellow boxes guide the orientation towards a more rational perspective. As the timesteps progress from $m = 0$ to $m = 3$, the scene quality is optimized from a coarse representation to a fine depiction, enriched with more semantic details.}
  \label{fig:experiment_layout}
\end{figure}

\paragraph{Effectiveness of ITFS}
As illustrated in \cref{fig:experiment_layout}, our method leverages implicit layout knowledge obtained from a 2D generation model to further optimize the 3D scene. The proposed ITFS strategy significantly enhances scene generation by refining object orientation and detail. The coach labels in the two yellow boxes guide the orientation to achieve a more rational and coherent perspective. As the timesteps progress from $m = 0$ to $m = 3$, the scene quality is optimized from a coarse representation to a fine, detailed depiction. This progression not only improves geometric accuracy but also enriches the semantic details, resulting in a more realistic and contextually appropriate rendering.


\begin{table}[!t]
    \centering
    \setlength{\tabcolsep}{1.3mm}
    \small 
    \begin{tabular}{l|ccccc}
        \toprule
        \multirow{2}{*}{Method} & \multicolumn{2}{c|}{User Study} & \multicolumn{3}{c}{Quality} \\
        \cmidrule{2-6}
        & $\uparrow$ Rationality & $\uparrow$ Quality & $\uparrow$ AS & $\uparrow$ IR  & $\uparrow$ CS $_{\%}$ \\
        \midrule
        Set-the-scene & 1.13 & 2.80 & 4.83 & 41.10 & 26.05 \\
        GALA3D & 2.61 & 3.22 & 5.07 & 42.67 & 25.40 \\
        DreamScene & 3.10 & 3.15 & 5.13 & 43.85 & 26.77 \\
        \textbf{Ours} & \textbf{4.10} & \textbf{3.86} & \textbf{5.31} & \textbf{48.83} & \textbf{28.44} \\
        \bottomrule
    \end{tabular}
    \caption{Quantitative evaluation across open soucre method on Indoor Scene Generation. Our method demonstrates consistent superiority across quality metrics (Rationality, Quality) and user study metrics (AS, IR, CS).}
    \label{tab:text_3d_campare}
\end{table}

\subsection{Quantitative Results}
\paragraph{User Study for Semantic Alignment} 
We conducted a user study to compare our approach with existing open source methods, focusing on scene quality and scene authenticity. We selected five common indoor scenes,~\ie~living room, kitchen, dining room, bedroom, and workplace, for quantitative evaluation. Thirty participants, including professional interior designers, Embodied-AI algorithm engineers, and graduate students, evaluated these scenes on a 1-to-5 scale, indicating their satisfaction from low to high. As shown in~\cref{tab:text_3d_campare}, our method achieved the highest scores across different downstream fields.

\paragraph{3D Scene Generation Quality}
As shown in~\cref{tab:text_3d_campare}, we tested our approach on three qualitative metrics. Aesthe. Score (AS)~\cite{ava} that evaluates aesthetic quality of scene. We also compared 3D generation quality on the ImageReward (T3Bench) (IR)~\cite{imagereward}, which refers to image reward applicantion in T3Bench, and CLIP Score (CS)~\cite{clipscore} to evaluation the text-3D scene alignment. Set-the-Scene~\cite{setthescene}, GALA3D~\cite{GALA3D} and DreamScene~\cite{dreamscene} both lack of scene diversity and rendering realism issues, bringing negatively increasing their overall scores. Our method performs well on all metrics, which is consistent with the results shown in the user study, which once again proves the superiority of our paradigm combining LLM, scene plan and ITFS generation.

%% file: sec/5_conclusion.tex
\section{Conclusion}
We present~\Ourmodel, an automatic framework for text-to-3D scene generation. Following a pipeline of ‘Reasoning-Grounding, Arrangement, and Optimization,’ our method enhances diversity, complexity, and efficiency for text-to-3D scene generation. By integrating LLMs for scene layout guidance, collision and reachability constraints for layout arrangement, and specific sampling strategies for rendering, our method generates explicit layout criteria to guide rational spatial placements. Experimental results demonstrate that~\Ourmodel~outperforms existing methods in generating diverse and complex scenes while requiring fewer computational resources for realistic 3D representation. 

\paragraph{Limitation}
Despite its computational efficiency, our method operates on an RTX 4090 GPU with 24GB VRAM. To balance efficiency and performance within this memory constraint, we limit the increase in the number of 3D Gaussians during later optimization stages. This results in insufficient resources to adequately model the fine-grained scattering behavior of light on complex materials and structures. Future work will focus on enhancing visual realism by addressing these limitations, particularly in improving the modeling of complex light transport and recovering finer geometric details, to generate scenes suitable for a broader range of demanding applications.


%% file: sec/6_appendix.tex
\input{appendix/high_level_prompt}
\input{appendix/high_level_prompt2}
\input{appendix/llm_gounding}

\section{Hierarchical LLM Agents Planning}
This section provides details about our five prompt templates used for LLM agents to generate long-term descriptions. Each module enforces physical plausibility constraints and syntactic standardization, allowing LLM agents to perform reasoning and grounding.

\paragraph{Floor and Wall Height Module}
This module structures room layout generation through explicit geometric and material constraints, as shown in Figure~\ref{fig:prompt-1}.
Rectangular boundary enforcement,~\eg~``each room is a rectangle'', eliminates irregular shapes, ensuring compatibility with downstream systems.
Coordinate precision,~\eg~``units in meters'', allows direct conversion from metric to 3D and avoids unit ambiguity.
Non-overlap connectivity rule,~\eg~``connected, not overlapped'', mimics the construction principles of the real world and enforces physically plausible spatial relationships.
Size thresholds,~\eg~``3--8\,m side length, $\leq 48\,\text{m}^2$ area'', prevent unrealistic room proportions while accommodating standard residential dimensions.

\paragraph{Doorway Module}
This module regulates inter-room connectivity through parametric constraints and contextual style alignment, as shown in Figure~\ref{fig:prompt-2}.
Typology limitation,~\eg~``doorframe/doorway/open", eliminates architecturally invalid connections, which can reduce the output of hallucinations.
Dimensional catalog,~\eg~``1m/2m enforces the compliance of the building code.
Style binding rule,~\eg~``complements room design", ensures material coherence. It can prevent contradictions, such as ``industrial metal door" connecting to ``traditional Japanese tatami room".

\paragraph{Window Module}
This module optimizes window design through constrained parametric selection and context-aware placement rules, as shown in Figure~\ref{fig:prompt-3}.
The type-size catalog (fixed/hung/slider with standardized dimensions) ensures the manufacture of specifications, eliminating invalid configurations in the LLM outputs.
Unified style constraints,~\eg~``within the same room, all windows must be the same type and size", maintain architectural coherence while reducing combinatorial complexity.
Base height parameter,~\eg~``50cm--120cm'' aligns with human anthropometry standards, preventing implausible placements such as floor-level windows in bathrooms.

\input{appendix/scene_style_extend}

\paragraph{Object Selection Module}
This prompt structures scene furnishing through constraint systems that enforce physical plausibility and realism, as shown in Figure~\ref{fig:prompt-4}.
Spatial hierarchy constraint,~\eg~``small objects on top'' optimizes space utilization while avoiding visual monotony or clutter.
Isolation constraint,~\eg~``Do not provide rug/mat, windows, doors, curtains, and ceiling objects,~\etc'', eliminates cross-module conflicts and prevents duplicate assets.
Dual-stage output structuring constraint,~\eg~``first use natural language to explain high-level design strategy, then follow \textit{.json} file format'', ensures schema consistency.

\paragraph{Object Alignment Module}
This module enables the generation of vivid architectural scenes and eliminates common LLM hallucination patterns, as shown in Figure~\ref{fig:prompt-5}.
Material anchoring requirement,~\eg~``clearly mention floor and wall materials", eliminates ambiguous descriptions such as ``nice floors" in the LLM output.
Structural purity constraint,~\eg~``no people/objects except architectural features", restricts the semantic space and allows focus on consistency of the wall, windows, and posters.

\input{appendix/code}

\begin{listing}[t!]
  \caption{Rendering Parameters}
  \label{lst:setting} 
  \begin{lstlisting}[style=mysnippet]
    # total training steps
    iters: 1500                
    densification_interval: 100
    density_start_iter: 100
    density_end_iter: 1500
    # Batch size
    batch_size: 1
    # Learning rates
    position_lr_init: 0.0008
    position_lr_final: 2.5e-05
    position_lr_delay_mult: 0.01
    position_lr_max_steps: 1200
    feature_lr: 0.01
    feature_lr_final: 0.005
    opacity_lr: 0.01
    opacity_reset_interval: 30000
    scaling_lr: 0.005
    scaling_lr_final: 0.005
    rotation_lr: 0.001
    rotation_lr_final: 0.0004
    texture_lr: 0.3
    geom_lr: 0.0002
    # Point cloud growth / sparsification
    max_point_number: 3000000
    percent_dense: 0.01
    densify_grad_threshold: 0.001
    density_thresh: 1.0
    # Camera and viewing parameters
    # vertical FOV range (deg)
    fovy: [76.0,96.0]
    elevation: 0
    # camera orbit radius range
    radius:[1.5,2.5]
    # yaw range (deg)
    ver: [-50,50]
    # File paths and resources
    mesh_format: glb
    outdir: null
    plyload: null
    # Prompt settings
    prompt: ""
    # Scene statistics
    room_size: ""
    room_num:
    assets_num:
    # Object size list with prompts
    object_size_list:
      - object: 1
        object_size:
        prompt: ""
    # Miscellaneous flags
    save: true
    sh_degree: 3

  \end{lstlisting}
\end{listing}

\section{AnyReach Camera Trajectory Sampling}
In this section, we present a detailed demonstration of how AnyReach camera trajectory sampling is implemented during the rendering stage.

\paragraph{Rendering Parameters}
In Listing~\ref{lst:setting}, we provide parameter values that are directly employed in our method, as referenced throughout the experimental implementation.

\begin{figure}[t!]
  \centering  
  \includegraphics[width=0.9\columnwidth]{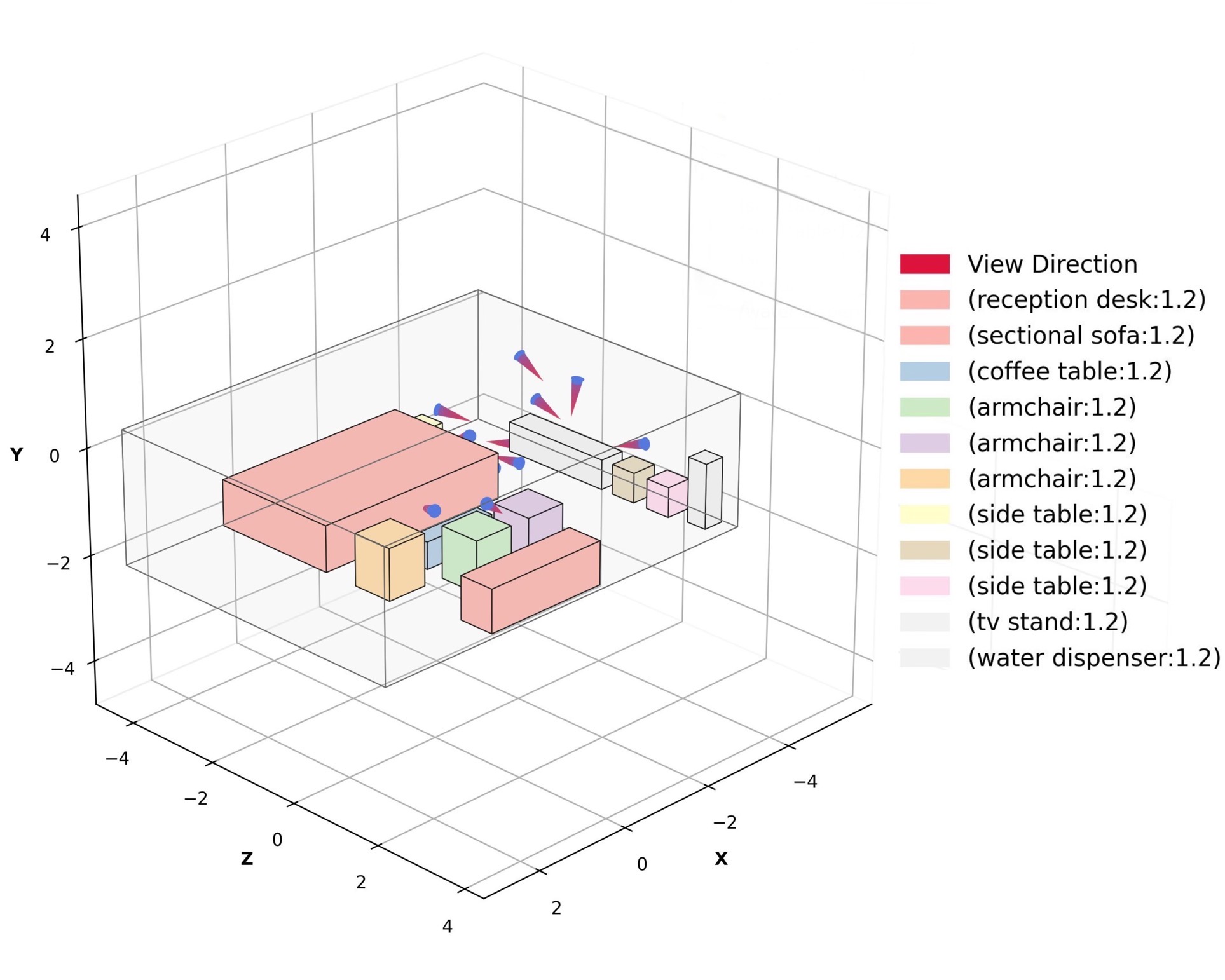}
  \caption{Camera trajectory sampling of \textbf{zoom-in} mode.}
  \label{fig:walk}
\end{figure}

\paragraph{Zoom-in Mode Trajectory}
Figure~\ref{fig:walk} demonstrates our ``zoom-in'' camera sample strategy generated using grid-based spatial reasoning. The sampling mode first discretizes the room into an occupancy grid, where cells are marked as obstructed (\eg~by furniture or wall) or navigable. Starting from the doorway, it plans sequential paths to key object centers using the A* algorithm, ensuring human collision-free movement through the room. The camera is positioned at the height of the human eye with a randomized elevation, while its orientation remains fixed to a central room target. The resulting trajectory simulates natural human movement in the real world, prioritizing the acquisition of local object-level details.

\begin{figure}[t!]
  \centering  
  \includegraphics[width=0.9\columnwidth]{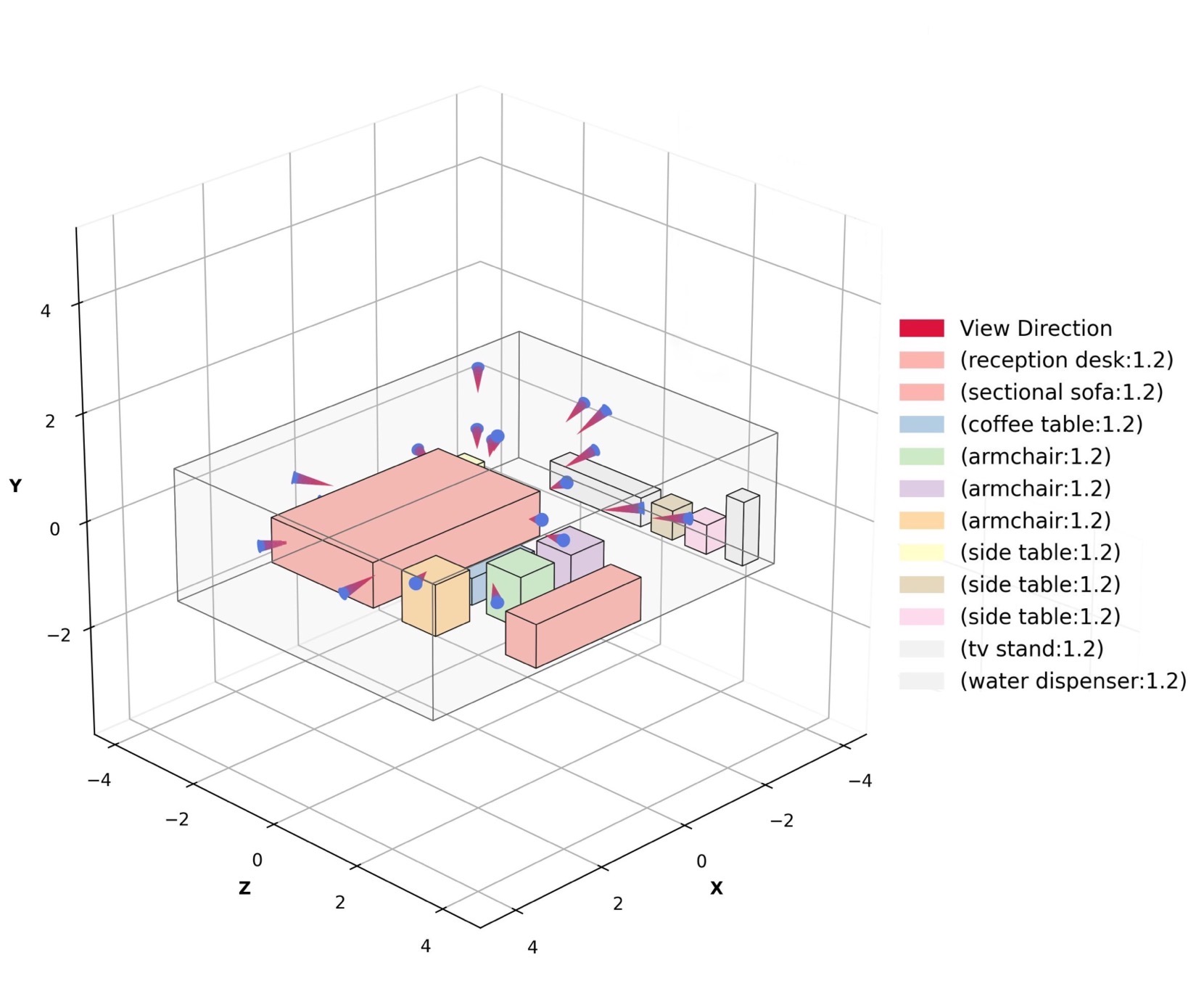}
  \caption{Camera trajectory sampling of \textbf{zoom-out} mode.}
  \label{fig:orbit}
\end{figure}

\paragraph{Zoom-out Mode Trajectory}
Figure~\ref{fig:orbit} shows the ``Zoom-out" spherical camera trajectory mode. It utilizes a spiral camera path, providing global observation of the background and objects within the scene through angular variations in azimuth and elevation. The system dynamically selects observation targets as the closest object center to each camera position. Each camera is directed toward the object's center through geometric alignment calculations, creating systematic inspection orbits. This approach generates comprehensive coverage of both background context and object layouts while maintaining collision-aware positioning within valid room boundaries.

\begin{figure}[t!]
  \centering  
  \includegraphics[width=0.9\columnwidth]{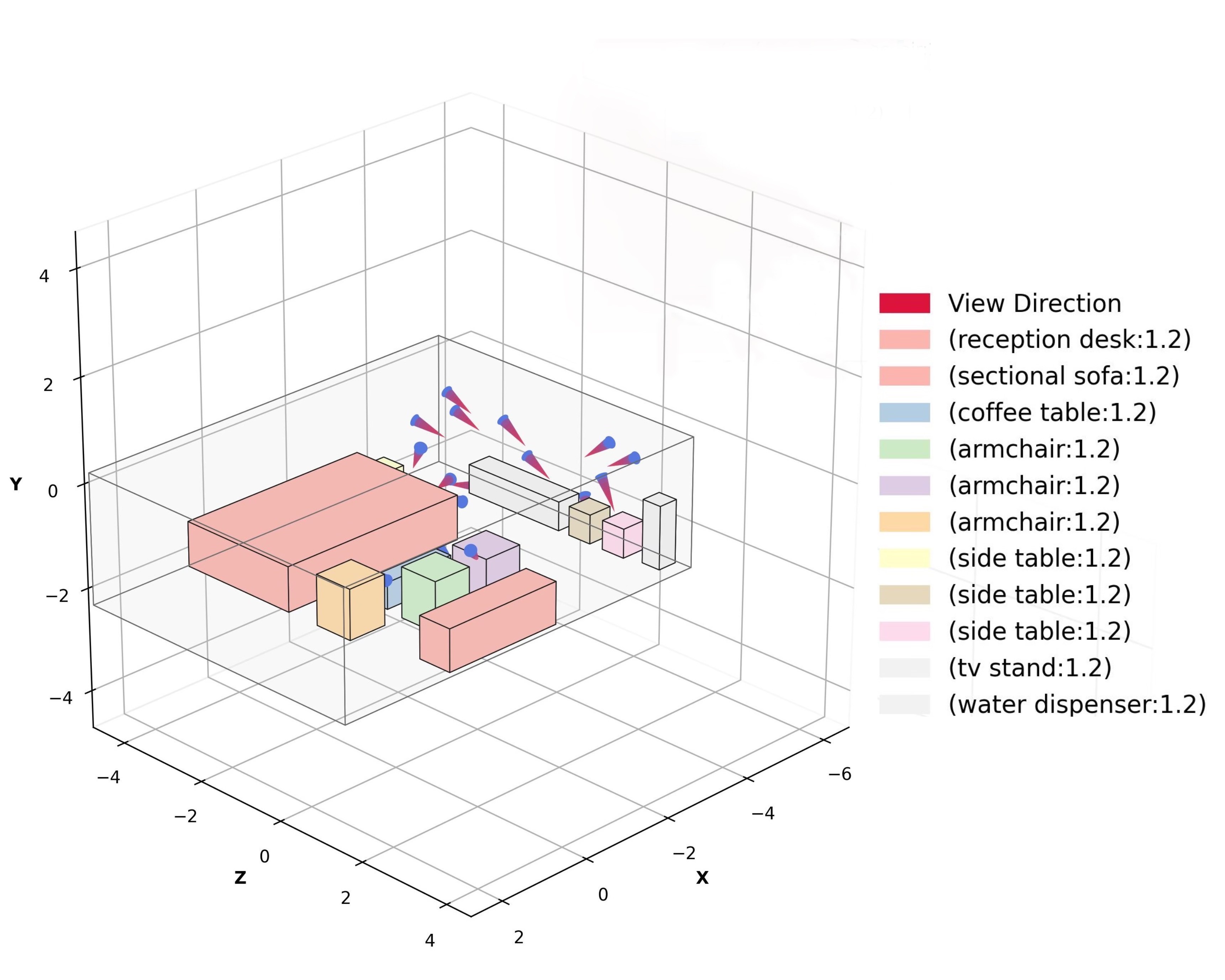}
  \caption{Camera trajectory sampling of \textbf{hybrid} mode.}
  \label{fig:mix}
\end{figure}

\paragraph{Hybrid Mode Camera Trajectory}
Figure~\ref{fig:mix} shows our hybrid sampling alternates between spiral global observation (zoom-out) and A* optimized object approach (zoom-in), creating a dynamic ``observe-approach-reset" cycle. This pattern minimizes idle transit time compared to sequential sampling while maintaining collision awareness through continuous occupancy grid validation during all transitions.

\begin{figure}[t!]
  \centering
\includegraphics[width=1\columnwidth]{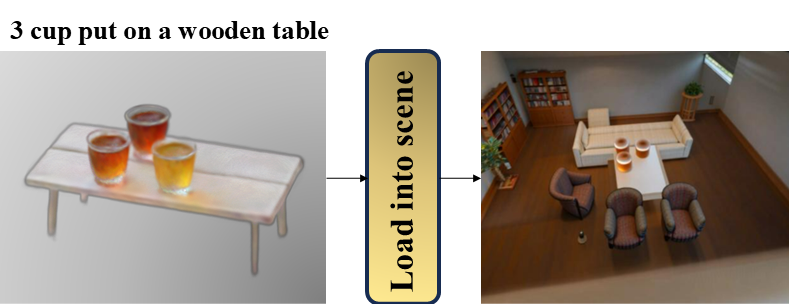}
  \caption{ Visualization of editing functionality for additional items.}
  \label{fig:edition}
\end{figure}

\begin{figure*}[t!]
  \centering  
  \includegraphics[width=1\textwidth]{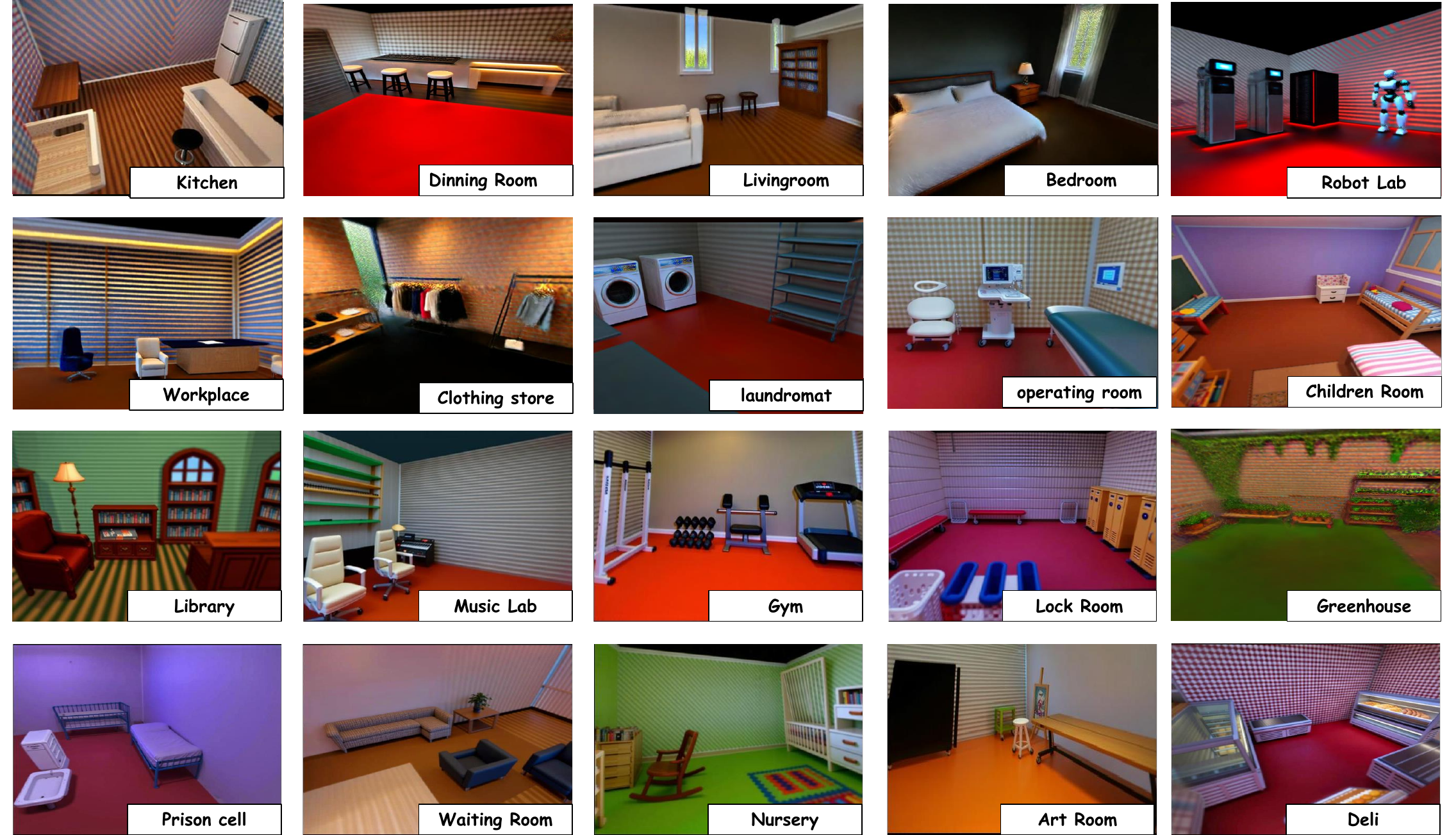}
  \caption{A diverse set of 3D scenes generated by~\Ourmodel, demonstrating its ability to create both common and uncommon room types.}
  \label{fig:diversity}
\end{figure*}

\section{Editing Functionality}
RoomPlanner supports the suitable placement of new objects (\eg~``a table with three cups") into existing 3D scenes. It automatically resolves spatial conflicts while maintaining realistic proportions and physical coherence. As shown in Figure~\ref{fig:edition}, the added cups are adapted to the scale of the scene and adhere to the rules of placement in the real world without manual intervention.


\section{Diversity of Generated Scenes}

In Figure~\ref{fig:diversity}, we present 25 additional scenes generated by~\Ourmodel. These examples encompass a wide range of room types, including common spaces such as kitchens, dining rooms, living rooms, and bedrooms, as well as less typical environments like robot labs, music labs, prison cells, and art rooms. This diverse selection highlights capability of ~\Ourmodel~to generate realistic and varied 3D interiors across both everyday and uncommon scenarios.

%% file: appendix/high_level_prompt.tex
\begin{figure}[b!]
    \begin{center}
    \begin{tcolorbox} [top=2pt,bottom=2pt, width=\linewidth, boxrule=1pt]
    {\footnotesize {\fontfamily{zi4}\selectfont
    \textbf{Floor plan Prompt:}
    You are an experienced room designer. Please assist me in crafting a floor plan. Each room is a rectangle. You need to define the four coordinates and specify an appropriate design scheme, including each room's color, material, and texture.
    Assume the wall thickness is zero. Please ensure that all rooms are connected, not overlapped, and do not contain each other.
    \begin{itemize}[leftmargin=10pt,itemsep=0.5ex,topsep=0.5ex]
            
        \item Note: the units for the coordinates are meters.
    
        \item For example:
        living room | maple hardwood, matte | light grey drywall, smooth | [(0, 0), (0, 8), (5, 8), (5, 0)]
        kitchen | white hex tile, glossy | light grey drywall, smooth | [(5, 0), (5, 5), (8, 5), (8, 0)]
        \item A room's size range (length or width) is 3m to 8m. The maximum area of a room is 48 m$^2$. Please provide a floor plan within this range and ensure the room is not too small or too large.
        \item It is okay to have one room in the floor plan if you think it is reasonable.
        \item The room name should be unique.
        
        Now, I need a design for \{input\}.
        Additional requirements: \{additional requirements\}.
        Your response should be direct and without additional text at the beginning or end.
    \end{itemize}
    \par}}
    \end{tcolorbox}
    \begin{tcolorbox} [top=2pt,bottom=2pt, width=\linewidth, boxrule=1pt]
    {\footnotesize {\fontfamily{zi4}\selectfont
    \textbf{Wall Height Prompt:}
    I am now designing \{input\}. Please help me decide the wall height in meters.
    Answer with a number, for example, 3.0. Do not add additional text at the beginning or in the end.
    }
    \par}
    \end{tcolorbox}
    \end{center}
    \caption{Prompt templates for LLM reasoning to generate the floor and wall height module.}
    \label{fig:prompt-1}

\end{figure}

%% file: appendix/high_level_prompt2.tex
\begin{figure}[t!]
\centering
    \begin{center}
    \vspace{4.5em}
     \begin{tcolorbox} [top=2pt,bottom=2pt, width=\linewidth, boxrule=1pt]
    {\footnotesize {\fontfamily{zi4}\selectfont
    \textbf{Doorway Prompt:}
    I need assistance in designing the connections between rooms. The connections could be of three types: doorframe (no door installed), doorway (with a door), or open (no wall separating rooms). The sizes available for doorframes and doorways are single (1m wide) and double (2m wide). \\
    
    Ensure that the door style complements the design of the room. The output format should be: room 1 | room 2 | connection type | size | door style.
    For example: \\
    exterior | living room | doorway | double | dark brown metal door \\
    living room | kitchen | open | N/A | N/A \\
    living room | bedroom | doorway | single | wooden door with white frames \\
    
    The design under consideration is \{input\}, which includes these rooms: \{rooms\}. \\
    The length, width, and height of each room in meters are:
    \{room\_sizes\} \\
    Certain pairs of rooms share a wall: \{room\_pairs\}. There must be a door to the exterior. \\
    Adhere to these additional requirements \{additional\_requirements\}. \\
    
    Provide your response succinctly, without additional text at the beginning or end.
    }
    \par}
    \end{tcolorbox}
    \end{center}
    \caption{Prompt templates for LLM reasoning to generate the doorway module.}
    \label{fig:prompt-2}
\end{figure}

\begin{figure}[t!]
    \centering
    \begin{center}
    \begin{tcolorbox} [top=2pt,bottom=2pt, width=\linewidth, boxrule=1pt]
    {\footnotesize {\fontfamily{zi4}\selectfont
    \textbf{Window Prompt:}
    Guide me in designing the windows for each room. The window types are: fixed, hung, and slider.
    The available sizes (width x height in cm) are:
    fixed: (92, 120), (150, 92), (150, 120), (150, 180), (240, 120), (240, 180)
    hung: (87, 160), (96, 91), (120, 160), (130, 67), (130, 87), (130, 130)
    slider: (91, 92), (120, 61), (120, 91), (120, 120), (150, 92), (150, 120)
    
    Your task is to determine the appropriate type, size, and quantity of windows for each room, bearing in mind the room's design, dimensions, and function.
    
    Please format your suggestions as follows: room | wall direction | window type | size | quantity | window base height (cm from floor). For example:
    living room | west | fixed | (130, 130) | 1 | 50
    
    I am now designing {input}. The wall height is {wall height} cm. The walls available for window installation (direction, width in cm) in each room are:
    {walls}
    Please note: It is not mandatory to install windows on every available wall. Within the same room, all windows must be the same type and size.
    Also, adhere to these additional requirements: {additional requirements}.
    
    Provide a concise response, omitting any additional text at the beginning or end.
    }
    \par}
    \end{tcolorbox}
    \end{center}
    \caption{Prompt templates for LLM reasoning to generate the window module.}
    \label{fig:prompt-3}
\end{figure}

%% file: appendix/llm_gounding.tex
\begin{figure}[t!]
    \centering
    \begin{center}
    \begin{tcolorbox} [top=2pt,bottom=2pt, width=\linewidth, boxrule=1pt]
    {\footnotesize {\fontfamily{zi4}\selectfont
    \textbf{Object Grounding Prompt:}
    You are an experienced room designer, please assist me in selecting large 
    \textbf{floor/wall} objects and small objects on top of them to furnish the room. You need to select appropriate objects to satisfy the customer's requirements.
    You must provide a description and desired size for each object since I will use it to retrieve object. If multiple identical items are to be placed in the room, please indicate the quantity and variance type (same or varied).
    Present your recommendations in JSON format:
    {
      "sofa": {
        "description": "modern sectional, light grey sofa",
        "location": "floor",
        "size": [100, 80, 200],
        "quantity": 1,
        "variance  type": "same",
        "objects  on  top": [
          {"object  name": "news paper", "quantity": 2, "variance  type": "varied"},
          {"object  name": "pillow", "quantity": 2, "variance  type": "varied"},
          {"object  name": "mobile phone", "quantity": 1, "variance  type": "same"}
        ]
      },
      "tv stand": {
        "description": "a modern style TV stand",
        "location": "floor",
        "size": [200, 50, 50],
        "quantity": 1,
        "variance  type": "same",
        "objects  on  top": [
          {"object  name": "49 inch TV", "quantity": 1, "variance  type": "same"},
          {"object  name": "speaker", "quantity": 2, "variance  type": "same"},
          {"object  name": "remote control for TV", "quantity": 1, "variance  type": "same"}
        ]
      },
      "painting": {
        "description": "abstract painting",
        "location": "wall",
        "size": [100, 5, 100],
        "quantity": 2,
        "variance  type": "varied",
        "objects  on  top": []
      },
      "wall shelf": {
        "description": "a modern style wall shelf",
        "location": "wall",
        "size": [30, 50, 100],
        "quantity": 1,
        "variance  type": "same",
        "objects  on  top": [
          {"object  name": "small plant", "quantity": 2, "variance  type": "varied"},
          {"object  name": "coffee mug", "quantity": 2, "variance  type": "varied"},
          {"object  name": "book", "quantity": 5, "variance  type": "varied"}
        ]
      }
    }
    }

    Currently, the design in progress is \textbf{INPUT}, and we are working on the \textbf{ROOM  TYPE} with the size of ROOM  SIZE.
    Please also consider the following additional requirements: REQUIREMENTS.
    
    Here are some guidelines for you:
    
    1. Provide a reasonable type/style/quantity of objects for each room based on the room size to make the room not too crowded or empty.
    2. Do not provide rug/mat, windows, doors, curtains, and ceiling objects that have been installed for each room.
    
    3. I want more types of large objects and more types of small objects on top of the large objects to make the room look more vivid.
    Please first use natural language to explain your high-level design strategy for \textbf{ROOM  TYPE}, and then follow the desired JSON format \textbf{strictly} (do not add any additional text at the beginning or end).
    }
    \end{tcolorbox}
    \end{center}
    \caption{Prompt templates for LLM grounding in object selection.}
    \label{fig:prompt-4}
\end{figure}

%% file: appendix/scene_style_extend.tex
\begin{figure}[t!]
    \centering
    \begin{center}
    \begin{tcolorbox} [top=2pt,bottom=2pt, width=\linewidth, boxrule=1pt]
    {\footnotesize {\fontfamily{zi4}\selectfont
    \textbf{Object Grounding Prompt:}
    Generate a concise prompt (under 35 words) describing a \textit{description} featuring \textit{floor} and \textit{wall}, 
    creating a vivid and realistic atmosphere. Focus on architectural details. The prompt must:
        \begin{itemize}
            \item Clearly mention the floor and wall materials.
            \item Convey a vivid, realistic atmosphere.
            \item Contain no people or objects other than architectural features (like windows, walls, posters).
            \item Return only the plain prompt text — no quotation marks, markdown, headers, or explanations.
        \end{itemize}}}
    \end{tcolorbox}
    \end{center}
    \caption{Prompt templates for LLM grounding in object alignment.}
    \label{fig:prompt-5}
\end{figure}

%% file: appendix/code.tex
\lstdefinestyle{mysnippet}{
  language=Python,
  basicstyle=\ttfamily\small,
  numbers=left,
  numberstyle=\tiny\color{black!70},
  numbersep=6pt,
  commentstyle=\color{green!60!black},
  keywordstyle=\color{black},
  stringstyle=\color{blue!70!black},
  showstringspaces=false,
  xleftmargin=1.5em,
}

\newcommand{\AlgoCode}[3]{
  \begin{algorithm}[htbp]
    \caption{#2}
    \label{alg:#1}
    \begin{lstlisting}[style=mysnippet]
#3
    \end{lstlisting}
  \end{algorithm}
}